\documentclass[conference]{IEEEtran}

\usepackage{cite}
\usepackage{amsmath,amssymb,amsfonts}
\usepackage{mathabx}
\usepackage{algorithmic}
\usepackage{subcaption}
\usepackage{graphicx}
\usepackage{textcomp}
\usepackage{xcolor}
\usepackage{float} 

\def\BibTeX{{\rm B\kern-.05em{\sc i\kern-.025em b}\kern-.08em
    T\kern-.1667em\lower.7ex\hbox{E}\kern-.125emX}}

\newcommand{\set}[1]{\left[#1\right]}

\newcommand{\eqnref}[1]{(\ref{eqn:#1})}

\newcommand{\eqnlabel}[1]{\label{eqn:#1}}

\newcommand     {\paren}[1]{\left(#1\right)}
\newcommand{\abs}[1]{\left|#1\right|}
\newcommand{\curlb}[1]{\left\{#1\right\}}
\newcommand{\squareb}[1]{\left[#1\right]}

\newcommand{\norm}[1]{\left\|#1\right\|}
    
\begin{document}

\title{\Large ReFormer: Generating Radio Fakes for Data Augmentation
}
\author{
\IEEEauthorblockN{\normalsize Yagna Kaasaragadda, Silvija Kokalj-Filipovic}
\IEEEauthorblockA{\small Rowan University\\
\small\em kaasar57@rowan.edu, kokaljfilipovic@rowan.edu}}

\maketitle
\begin{abstract}
We present ReFormer, a generative AI (GAI) model that can efficiently generate
synthetic radio-frequency (RF) data, or RF fakes, statistically similar to the data it was trained on, or with modified statistics, in order to augment  datasets collected in real-world experiments.  For applications like this, adaptability and scalability are important issues. This is why ReFormer leverages transformer-based autoregressive generation, trained on learned discrete representations of RF signals. By using prompts, such GAI can be made to
generate the data  which complies with specific constraints or conditions, particularly useful for training channel estimation and modeling. It may also
leverage the data from a source system to generate training data
for a target system. We show how different transformer architectures and other design choices affect the quality of generated RF fakes, evaluated using metrics such as precision and recall, classification accuracy and signal constellation diagrams.   
\end{abstract}

\vspace{-2mm}
\section{Introduction}
Generative AI models (GAI) are the key technology for advancing modern wireless communication systems. They offer novel solutions for tasks such as signal generation, channel modeling, and system optimization.  The most popular among generative models for wireless communications are diffusion models, demonstrating superior performance in various applications.
Diffusion models are a class of probabilistic generative models that generate data by reversing a gradual noising process. Learning the iterative reverse process, called  "denoising", allows for generating signals starting from pure noise.
Researchers have proposed using denoising diffusion probabilistic models (DDPMs) to enhance communication systems with non-ideal transceivers and complex channels \cite{MIMOChannelEstimScore,  ScoreDiffChannel}. These models have shown significant improvements in bit error rates, especially in low signal-to-noise ratio conditions, by effectively denoising received signals. 
 DDPMs have been also used to probabilistically shape constellation symbols \cite{letafati2023generative}, optimizing the probability of symbol occurrence to enhance information rates and communication performance. This approach allows for adaptive signal design, aligning transmitted signals more closely with varying channel conditions. 
As the real-world radio frequency (RF)  data is scarce, GAI can be used to create large amounts of additional data for training machine learning models, including the ones mentioned above. Augmented data improves model performance in diverse scenarios. DDPMs are highly computationally inefficient for generative synthesis of large RF datasets, as the denoising process goes through iterations in both the training and the generation phase. 

Generative Adversarial Networks (GANs) have also been utilized for synthesizing realistic radio signals \cite{wang2023radioGAN} and channel estimation \cite{WGAN, zhou2024tiregan}. For instance, the "Radio GAN" framework \cite{wang2023radioGAN} employs an unrolled generator design combined with previously estimated pure signal distribution as a prior.  To synthesize high-quality radio signals, this method  learns transmitter characteristics and various channel effects through modeling the underlying sampling distribution. 
The drawback of this approach is that GANs are extremely difficult to train, which complicates any ablation analysis and adaptations.  
Given the complexity and diversity of multi-antenna systems and other Next-Generation network (NextG) scenarios, replacing diffusion and GAN-based GAI models by  transformer-based architectures may be a good solution for scalability in  generating RF fakes. Adaptation through prompting is another attractive feature. In this paper we present a simple and tractable generative model named ReFormer, which creates  RF signal fakes sampled from the learned distribution of the training signals captured by a decoder-only transformer. Section ~\ref{sec:system} introduces  deep learning architectures,  methods and data utilized to train the ReFormer GAI. Section ~\ref{sec:eval} describes methods of evaluation. Section ~\ref{sec:res} presents results of evaluation. We show how different transformer architectures and other design choices affect the quality of generated RF fakes, evaluated using metrics such as precision and recall, classification accuracy and signal constellation diagrams. 
%%%%%%%%%%%%%%%
\vspace{-1mm}
\section{System Model}\label{sec:system}
We aim to train a generative model to produce RF signals $X_g$ whose probability distribution $P(X_g)$ closely matches  probability distribution of the RF datapoints presented in the training dataset. In the big picture, this will allow us to later include the promts $C,$ which characterize desired variations in $X_g$ according to a conditional distribution $P(X_g|C).$ Such an ability will be essential for generating massive amounts of diverse datapoints for the augmentation of RF datasets, which is very important in the field of RF machine learning given the difficulties in collecting this type of data by sensing the RF spectrum.
\vspace{-1mm}
\subsection{Data}\label{subsec:torchsigdata}
%\vspace{-1mm}
Our dataset comprises of datapoints which represent sampled RF signals, of the types used in modern wireless communications. Each datapoint arises from a specifically modulated sequence of random information bits, converted to a baseband RF signal. The datapoint $x$ can be represented as $x=\squareb{Re_i+j Im_i}, i = 1\cdots p$ with $j=\sqrt{-1}$.
Here, a modulated signal
$u$ is obtained as $u = M_s(b)$, where $s \in \mathcal{S}$ is the employed
modulation scheme, with $S$ denoting the finite set of available
digital modulation schemes. In this paper, $$\mathcal{S} = \squareb{4ask,8pam,16psk,32qam-cross,2fsk,ofdm256}.$$ For any $s$, $M_s = \curlb{0,1}^m \rightarrow \mathcal{C}^n$ describes the modulation function of modulation class $s$. The random sequence of bits $b=\curlb{0,1}^m$ of length $m$ is encoded into a sequence of complex valued numbers of length $n,$ where the complex sample $c_i=Re_i +j Im_i,1 \leq i \leq n$ encodes the modulation phase $\phi=\arctan{Re_i/Im_i},$ and amplitude $a_i = \sqrt{Re_i^2+Im_i^2}.$ We create datapoints as sub-sequences  $x$ of $u \in \mathcal{C}_n,$ of length $p = 1024.$ We prepared the training dataset $X_{train}$ by using an open-source library {\em torchsig} featured in \cite{torchsig}. $X_{train}$ contains RF samples of high SNR, for both simplicity and  future utility in creating various signal augmentations controlled by a prompt. The torchsig library function {\em ComplexTo2D} is used to transform vectors of complex-valued numbers into  2-channel datapoints, traditionally referred to as {\em I} and {\em Q} components. Each channel is comprised of $p$ real numbers, previously normalized. Channel 1 contains real components $I\in \mathbb{R}^p$ and channel 2 contains the imaginary ones $Q\in \mathbb{R}^p.$  Depending on the modulation, $x$ contains more or less mappings of the original random sequence of bits $b.$ $\mathcal{S}$ contains diverse modulation orders (how many original bits are represented by a single complex value). By using $p > 1024,$ we may be able to better train the proposed generative model. However, this would require more complex neural net architecture and longer training. Future work will explore the effect of $p$ on the quality of the generated RF fakes. 
\subsection{Method}
Prior work \cite{BalkanCom24hqarf, Rodriguez2024DeepLearnedCF} used a hierarchical VQVAE model to learn several posterior distributions, including $P(Z_q|x)$ in its first level of hierarchy,
%\begin{align}
%P(Z_q|x) &= \eX^{-\norm{Z_e(x) - %q_k}^2}, \eqnlabel{post}
%\end{align}
where $Z_q$ is the vector-quantized version of the latent $Z_e$ mapped from the RF datapoint $x$ by the trained encoder. Now we want to learn the prior $P(x),$ which will allow us to sample $P(x)$ and generate similar RF fakes. However, to decrease the training complexity of the model which parameterizes this distribution, we choose to leverage $P(Z_q|x)$ for mapping $X_{train}$ to a discrete space and than learn the prior of $Z_q.$ Generating the fakes in the discrete latent domain is easier, and we can directly leverage the power of the transformer architecture \cite{deconly}. Once we obtain a fake $\hat{Z}_Q,$ we will use the VQVAE decoder to map it back to the original space, resulting in a fake RF datapoint $x_g.$  
%In the hiearchical VQVAE in \cite{}, %$P(Z_q|x)=P(Z^0_q|x)$ , while we also %learned $P(Z^{i}_q|P(Z^{i-1}_q)$
\subsection{Learning Discrete Posterior}
Using a Vector Quantized 
Variational Autoencoder model \cite{vqvae2017neural}  
 to learn an 
efficient, discrete, low-dimensional 
representation $Z_Q$ of every RF datapoint $x,$ allows us to create a discrete mapping of $X_{train}.$ The dataset $ZD$ is mapped from the training dataset $X_{train},$ using the mapping $\Phi_Q(E_\Theta()): X_{train}\rightarrow ZD,$ where $E_\Theta(x), x\in X_{train}$ is the trained VQVAE encoder with parameters $\Theta,$ and $\Phi_Q(Z)$ is the  vector quantizer based on the trained codebook $Q.$ Equations~\eqnref{vqvaefwd}  represent the trained VQVAE model utilized for the above mapping.
%%%%%%%%%%%%%%%%%%%%
\begin{align}
\vspace{-1mm}
\nonumber z &= E(x, \theta_E) \\
\nonumber z_q &= \Phi(z, Q, \theta_Q) \\
    \hat{x} &= D(z_q, \theta_D) \eqnlabel{vqvaefwd}
\end{align}
%%%%%%%%%%%%%%%%%%%%
Subsection~\ref{subsec:discprior} will show how we learn the prior $P(Z_Q)$ of the discrete latent sequences generated from $X_{train}$: we learn $P(Z_Q)$ probability model by training the transformer to learn its autoregressive form $P(Z^i_Q|Z^{1\cdots (i-1)}_Q)$. 
%%%%
Our VQVAE model has three components: (1) an 
encoder block $E: \mathbb{R}^{2\times1024}\xrightarrow{}\mathbb{R}^{\ell\times d_s}$  
which down-samples the input by a 
factor of 2 in each convolutional channel 
dimension and produces an output $z$ such 
that the number of channels of the 
output is equal to the codebook dimension $\ell=64$; we refer to each $z$ column-partition of size $\paren{64\times1}$ as a slice $z_i,$ resulting in $d_s=512$ slices (2) a decoder block $D: \mathbb{R}^{\ell\times d_s}\xrightarrow{}\mathbb{R}^{2\times1024}.$  (3) a vector quantizer $\Phi$, 
which is applied to the output of $E$
amd provides an output to be decoded by $D.$ 

Vector 
quantization (VQ) is a process which 
discretizes the latent space (output of $E$). Each latent slice is   discretized by applying the quantization function $ \Phi $, which maps it to an element of $ Q  = \{ e_i \mid 0 \leq i < N\},$ consisting of $N$ codewords, the learnable embedding vectors of the same length ${\ell}$.  VQ is effectively mapping each slice
to an index of the ordered set of codewords, referred to as the
codebook $Q.$
 The VQ-VAE model aims to  
learn the optimal codebook vectors through the 
training process. The decoder block, denoted $ D(z_q) $, is tasked with reconstructing the original input $x$ from the mapped codebook vectors. While the most intuitive approach for $ \Phi $ is to simply perform a nearest neighbor lookup, more advanced techniques such as stochastic codeword selection $\Phi_{SC}$ \cite{StochQ} or heuristics such as exponential moving averages (EMA) $\Phi_{EMA}$ \cite{vqvae2} can be employed instead. We here present the results obtained using stochastic mapping $\Phi_{SC}.$ In the equations~\eqnref{vqvaefwd}, $z$ is the latent representation of $x$ at the output of $E$ while $z_q$ is its quantized version. 

\begin{figure}[h]
\vspace{-2mm}
\centering
\hspace{-1mm}\includegraphics[width=0.48\textwidth, height = 4.4cm]{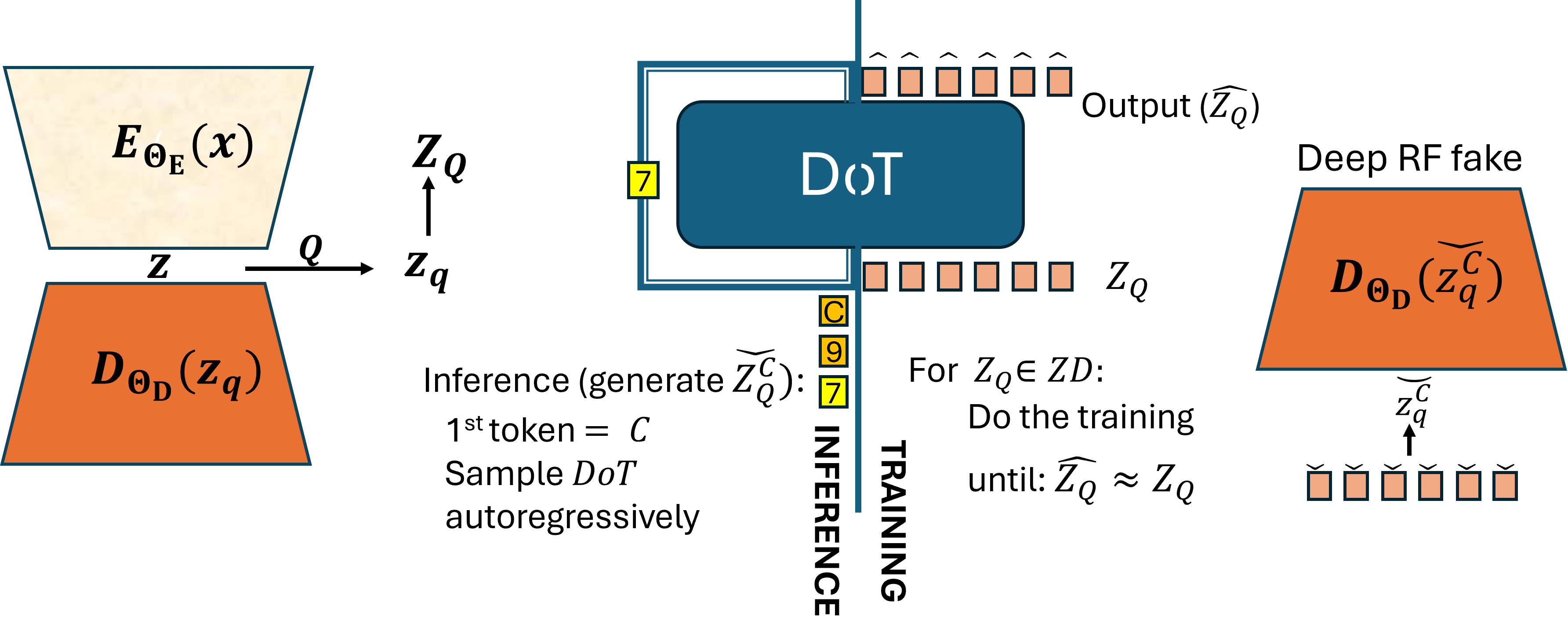}
\vspace{-2mm}
\caption{ReFormer approach: VQVAE latent token sequences $Z_Q$ are used to train decoder-only transformer (DoT), making it capable to generate fake sequences $\widecheck{Z_Q},$ which are turned into RF fakes via $D(\widecheck{z_q}).$
}\vspace{-3mm}
\label{fig:refall}
\end{figure}

We are interested in another form of the quantized latent $z_q$, denoted in Fig.\ref{fig:refall} by $Z_Q,$ which replaces the codewords quantizing the $z$ slices with their indices in the codebook. We refer to those indices as tokens $Z_Q^j \in \curlb{0,\cdots N-1}, N=\abs{Q}=128,$  where $j$ is the slice indices $j\in \curlb{1, \cdots,d_s}.$
%%%%%%%%%%%%%%%%%%%%.
\subsubsection{Training VQVAE}
The ability of the VQVAE to faithfully reconstruct an input $x$ will suffer a great degree if the codebook is 
not trained properly \cite{huh2023straightening}, leading to mode collapse. 
Our objective for this model is for 
the reconstructed output, referred to as $\hat{x}$, 
to be as close as possible to the input (or 
the original) $x$. Thus, we use reconstruction between $x$  and $\hat{x}$:
$\mathcal{L}_{rec}=\frac{1}{p}\sum_{i=1}^{p}{\paren{x_i-\hat{x}_i}^2}$
We also need to 
incentivize proper training of the 
codebook. Therefore, we introduce 2 
other terms to our loss function, known as the 
quantization loss $\mathcal{L}_{quant}$ and the commit loss $\mathcal{L}_{commit}$. 
$\mathcal{L}_{quant}$ measures the 
degree to which the codebook should be 
trained with respect to the output of the 
encoder, while $\mathcal{L}_{commit}$  measures
the degree to which the encoder should 
be trained with respect to the 
codewords (Eqn.~\eqnref{commit}).
%%%%%%%%%%%%%%%%%%%%
\begin{align}
\nonumber \mathcal{L}_{quant}&=\left\|sg\set{E(x)}-\Phi\paren{E(x)}\right\|_2^2 \\
\mathcal{L}_{commit}&=\left\|E(x)-sg\set{\Phi\paren{E(x)}}\right\|_2^2, \eqnlabel{commit}
\end{align}
where $sg\set{\cdot}$  denotes the stop 
gradient function. $sg\set{\cdot}$ acts as the 
identity function during the forward pass, while in the 
backwards pass it produces a $0$-valued partial derivative with respect to 
all trainable parameters .
  
As the codebook is more difficult to train 
than the autoencoder, we
multiply $\mathcal{L}_{commit}$  by $0 < \beta < 1$
such that the codewords are "more trainable" 
than the  
encoder. As we apply stochastic quantization $\Phi_{SC}$ by sampling the codebook \cite{takida2022sq} according to the learned discrete posterior
\begin{equation}
P(Z_Q\set{i} = k|x) = e^{-\norm{z_i(x)-e_k}^2},\eqnlabel{post}
%= \eX^{-\norm{z_i(x)-e_k}^2}, 
\end{equation}
the loss also must include the KL divergence between the posterior in \eqnref{post}  and the discrete uniform prior $P_d(k)=1/N$, $KL(P(k|x)||P_d(q))$. Hence, the complete loss function is:
\begin{equation}
\mathcal{L}_{total}=\mathcal{L}_{rec}+\mathcal{L}_{quant}+\beta\paren{\mathcal{L}_{commit}+KL}.
\end{equation}
%More detailed description of the training process is given in \cite{BalkanCom24hqarf, Rodriguez2024DeepLearnedCF}.
\subsection{Learning Discrete Prior}\label{subsec:discprior}
To learn the prior in the discrete space, we train a decoder-only transformer (DoT) on the dataset $ZD,$  mapped from the training dataset $X_{train},$ using the mapping $\Phi_Q(E_\Theta()): X_{train}\rightarrow ZD.$  

    The DoT model \cite{deconly}, also known as the autoregressive transformer, is a popular generative model derived from the original encoder-decoder transformer model  \cite{vaswani2017attention}. The encoder stack is eliminated from this transformer, leaving the decoder stack to learn the causal attention structure in $Z_Q.$ Once trained, the model can perform autoregressive sequence generation. The  self-attention mechanism enables the model to predict the next token given all previous tokens. DoT can also include cross-attention, allowing for different creative methods of obtaining the cross-attending context.  The training relies on the cross-entropy loss (CE) between the masked indexed output of the VQ-VAE quantizer $Z_Q$ and its estimate $\hat{Z}_Q$ at the output of the transformer (see Fig.~\ref{fig:refall}). The latent fake generation, is an auto-regressive inference. It starts from a random token, and continues generating subsequent tokens of a fake $\hat{Z}_Q$ auto-regressively. We trained the transformer using a simple form of cross-attending context by preceding each datapoint $Z_Q \in ZD$ with its class token $C\in\curlb{0,\cdots, 5}.$ Hence, during the inference, to indicate what class of the latent fake we aim to generate, we start the autoregressive generation with the token $C.$ Once the latent fake  $\hat{Z}_Q$ is generated, we map it into its non-indexed version $\hat{z_q}$ and transform it by the VQ-VAE decoder into a RF fake.
%%%%%%%%%%%%%%%%%%%%%%%%%%%%%%
\subsection{Training the Transformer Models}\label{subsec:transformer_training}
To train the transformer model, we used two different architectures: the nano-GPT architecture \cite{nanogpt} and the MONAI transformer \cite{monai}. Both models were trained to learn the autoregressive representation of the discrete latent space $Z_Q$, which was generated by the VQ-VAE quantization process.

The discrete latent sequences $Z_Q$ contain $d_s$ tokens, where each token represents an index of the nearest codeword from the trained VQ-VAE codebook. To prepare the training data for the transformer, we constructed $ZD_{\text{X}}$ and $ZD_{\text{Y}}$ as follows:
\begin{itemize}
    \item $ZD_{\text{X}}$ consists of $d_s$ tokens, where the first token is a class label $C \in \{0, 1, \dots, 5\}$ representing the modulation scheme. The remaining $d_s - 1$ tokens are indices derived from the quantized $Z_Q$ sequence.
    \item $ZD_{\text{Y}}$ consists of $d_s - 1$ tokens, representing the indices from the second token onward in the $Z_Q$ sequence.
\end{itemize}

\subsubsection{Training Process}
The training process for both the GPT and MONAI transformer models is designed to minimize the CE loss between the predicted indices and the ground truth indices in $ZD_{\text{Y}}$. The steps involved in the training process are described below:
\begin{enumerate}
    \item \textbf{Input Preparation:} For each training example, the sequence $ZD_{\text{X}}$ is fed into the transformer model. The transformer generates a sequence of logits, where each logit corresponds to a prediction (out of N numbers) for the next token in the sequence.
    \item \textbf{Autoregressive Training:} The transformer is trained in an autoregressive manner. For a sequence $ZD_{\text{X}} = [C, t_1, t_2, \dots, t_{d_s - 1}]$:
    \begin{itemize}
        \item At step $n$, the model uses the first $n$ tokens of $ZD_{\text{X}}$ as input to predict the $(n+1)$-th token: i.e., given $[C, t_1, \dots, t_n]$, the model predicts the token $t_{n+1}$.
        \item. At step $d_s - 1,$ it stops.
    \end{itemize}
    \item \textbf{Loss Function:} The loss function is the CE loss between the predicted logits of the indices and the ground truth tokens in $ZD_{\text{Y}}$. Formally, the loss for a single example is computed as:
    \[
    \mathcal{L} = -\frac{1}{d_s - 1} \sum_{n=1}^{d_s - 1} \log P(z_{n+1}|C, z_1, z_2, \dots, z_n),
    \]
    where $P(z_{n+1}|C, z_1, z_2, \dots, z_n)$ is the predicted probability of the $(n+1)$-th token given the preceding $n$ tokens.
    \item \textbf{Inference and Loss Calculation:} During inference, the model generates tokens one by one in an autoregressive manner, starting with the class label $C$ and the random first token $z_1$. 
\end{enumerate}
%%%%%%%%%%%%%%%%%%%%%%%%%%%
\subsubsection{Implementation and Training Details}
We implemented the nano-GPT and MONAI transformer architectures using PyTorch. The transformers consist of multiple layers of self-attention and feedforward networks, with each layer designed to capture the dependencies between tokens in the sequence. The training was performed using a learning rate scheduler, with the Adam optimizer applied to minimize the cross-entropy loss.

Both models were trained for 100 epochs with a batch size of 32, using a maximum sequence length of $d_s=512$ tokens. We continued training both models  after the 100 epochs but the MONAI validation and training losses stated to diverge. Hence, the models that are evaluated here are those trained with 100 epochs each. This training enabled the transformer to learn an autoregressive mapping of the latent discrete space, providing a robust mechanism for generating new latent sequences $\widecheck{Z_Q}$ during inference. These latent sequences are then transformed back into the original RF signal space using the VQ-VAE decoder $\hat{x}=D(\widecheck{z_Q}, \theta_D).$
%%%%%%%%%%%%%%%%%%%%%%%%%%
\section{Methods of Evaluation}\label{sec:eval}
To evaluate the effectiveness of our approach, we utilized multiple evaluation techniques targeting both the VQVAE and the transformer-generated outputs. 
%%%%%%%%%%%%%%%%%%%%%
\subsection{Evaluation of VQVAE}
\begin{figure}[h]
\centering
\includegraphics[width=0.4\textwidth,height=3.5cm]{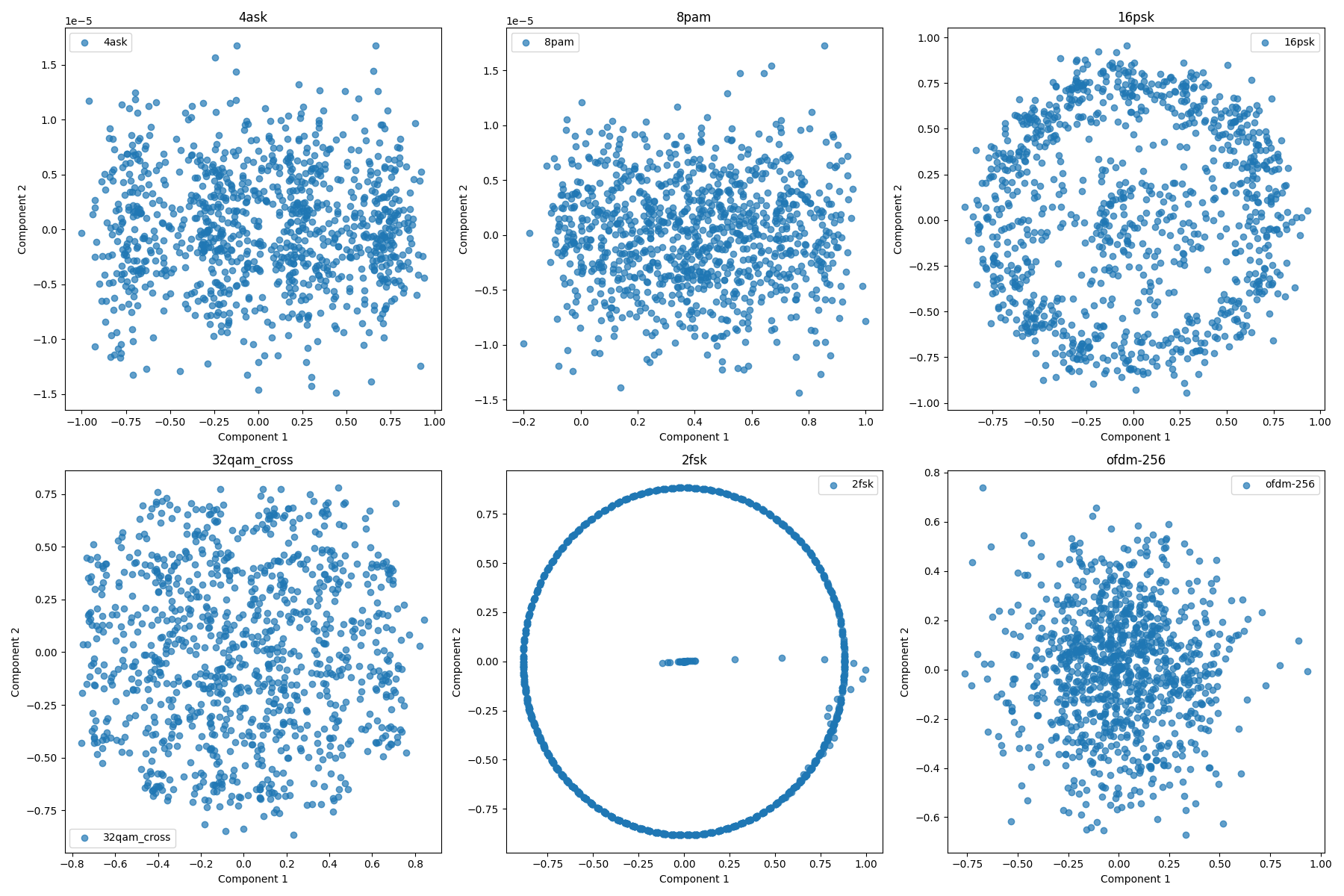} % Replace with actual file path
\caption{Original Signal Constellations.}
\label{fig:Original}
\end{figure}

\begin{figure}[h]
\centering
\includegraphics[width=0.4\textwidth,height=3.5cm]{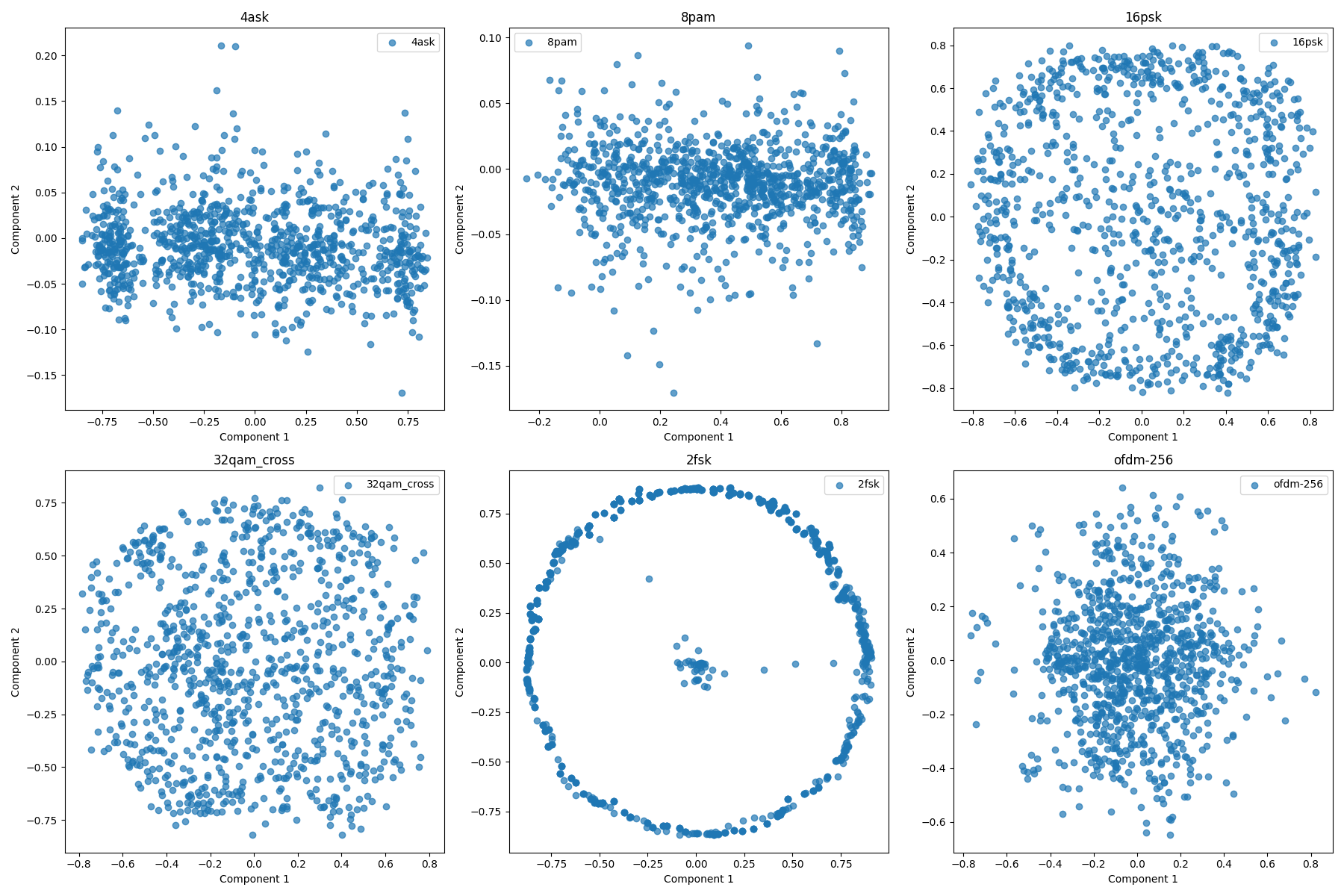} % Replace with actual file path
\caption{VQVAE Reconstructions Constellations.}
\label{fig:Reconstructions}
\end{figure}

\begin{figure*}
    \centering
    \
    \begin{subfigure}{0.2\textwidth}
        \centering
        \includegraphics[width=\textwidth]{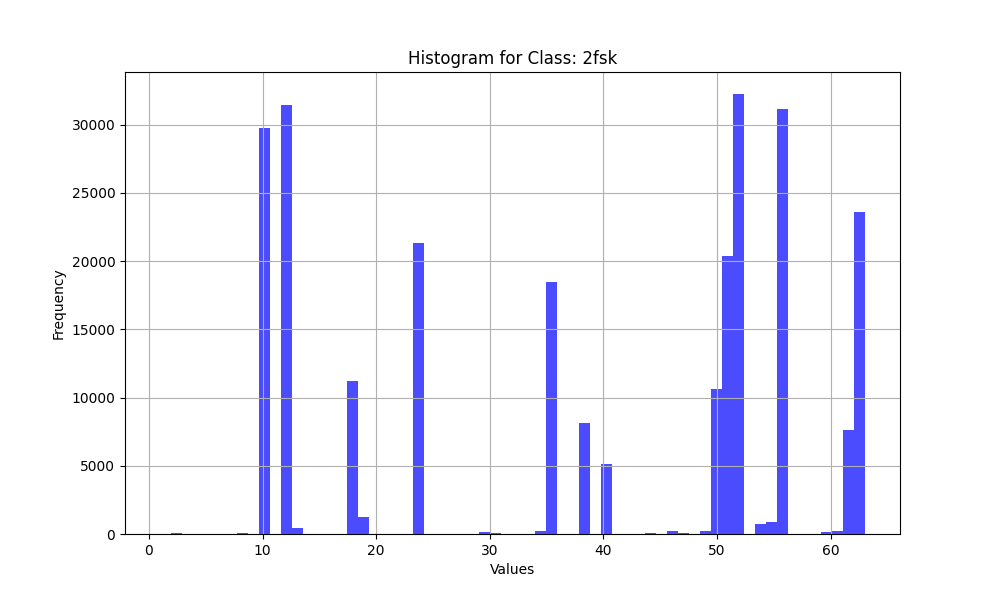}
        \caption{2FSK}
    \end{subfigure}
    \begin{subfigure}{0.2\textwidth}
        \centering
        \includegraphics[width=\textwidth]{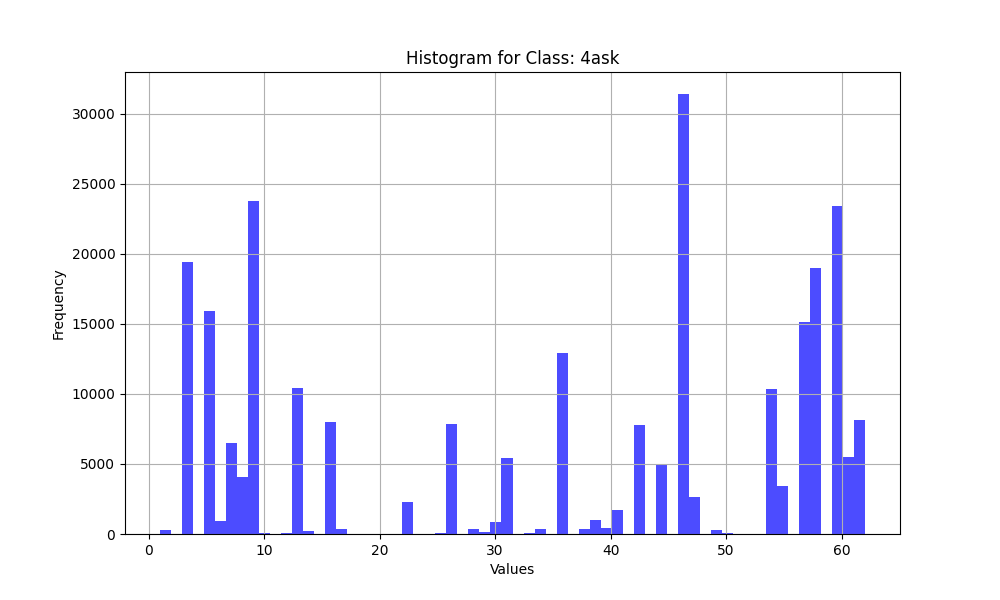}
        \caption{4ASK}
    \end{subfigure}
    \begin{subfigure}{0.2\textwidth}
        \centering
        \includegraphics[width=\textwidth]{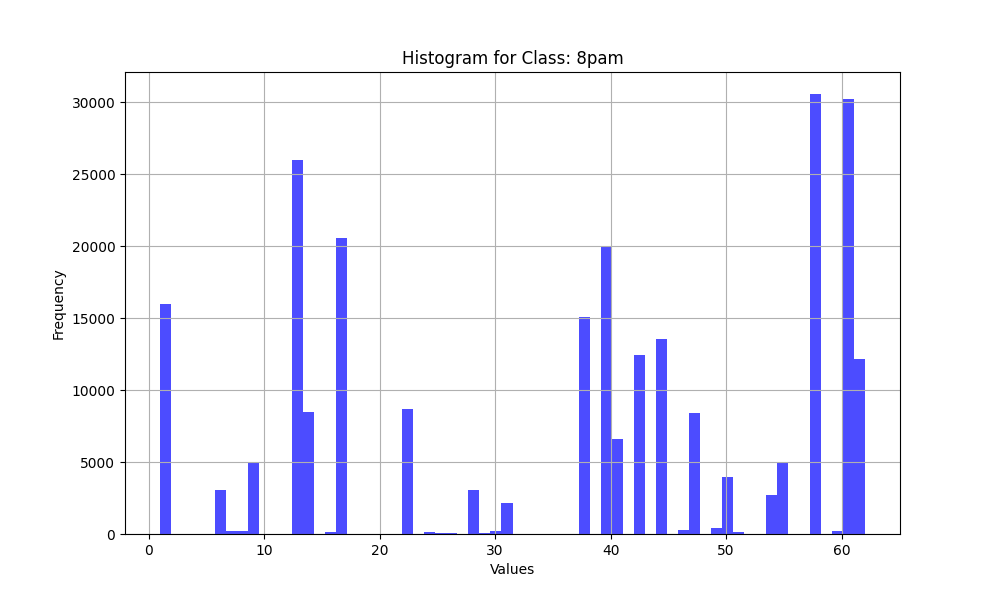}
        \caption{8PAM}
    \end{subfigure}
    \begin{subfigure}{0.2\textwidth}
        \centering
        \includegraphics[width=\textwidth]{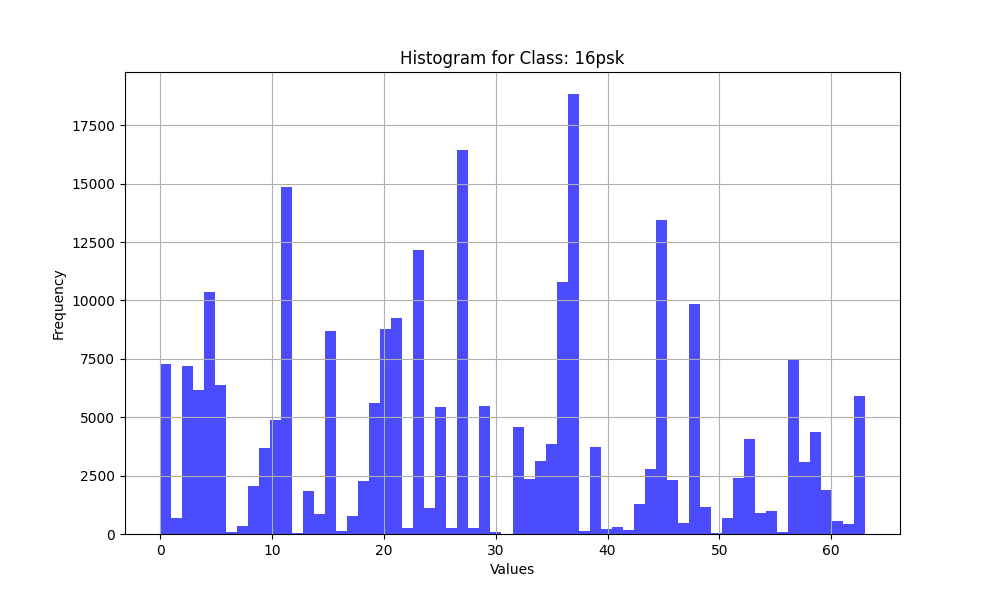}
        \caption{16PSK}
    \end{subfigure}
    \begin{subfigure}{0.2\textwidth}
        \centering
        \includegraphics[width=\textwidth]{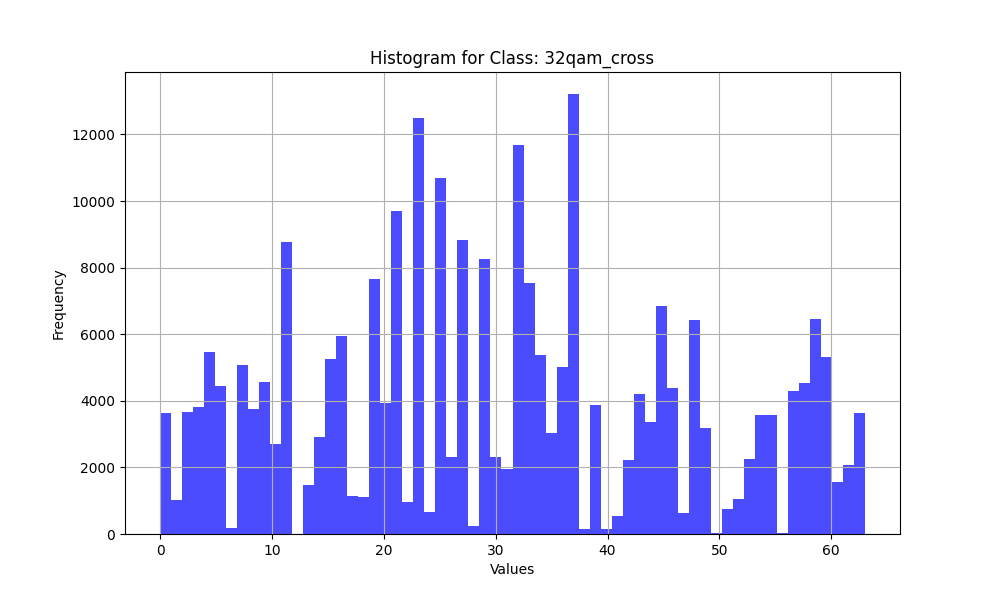}
        \caption{32QAM-Cross}
    \end{subfigure}
    \begin{subfigure}{0.2\textwidth}
        \centering
        \includegraphics[width=\textwidth]{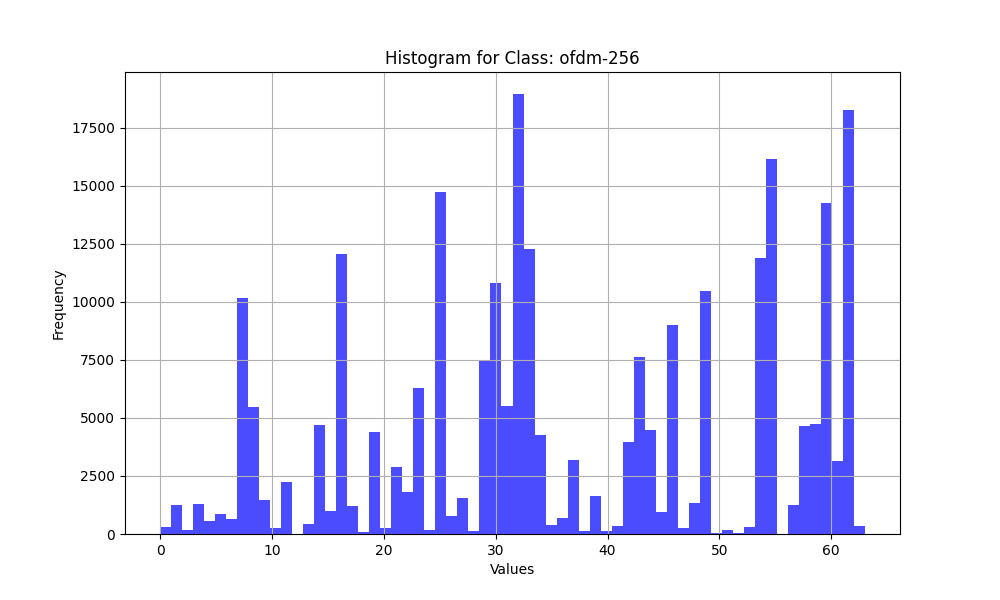}
        \caption{OFDM-256}
    \end{subfigure}
    \caption{Codebook usage histograms of reconstructions of VQVAE.}
    \label{fig:codebook_usage_VQVAE}
\end{figure*}

The VQVAE model was evaluated using a variety of reconstruction metrics.  Most importantly, we analyzed the codebook usage for each class (Figure \ref{fig:codebook_usage_VQVAE}). This  helps identify potential mode collapse, which would result in skewed histograms of codebook usage for one or more classes. Ensuring uniform codebook usage across classes validates proper training of the VQVAE and indicates robustness in the learned representations. Results of these evaluations are discussed in subsequent sections.
Additionally, we created 10 VQVAE reconstructions for each class and visualized the constellation diagrams for the I/Q samples. These diagrams (Fig.~\ref{fig:Reconstructions}) were compared to the original class constellations (Fig.~\ref{fig:Original}) to assess the fidelity of reconstructions visually. 
%%%%%%%%%%%%%%%%%%%%%%
\subsection{Fidelity and Diversity Evaluation}
For evaluating the fidelity and diversity of the generated fakes, we adopted the metrics described in \cite{kim2024topprrobustsupportestimation}, specifically Topological Precision and Recall (TopP\&R). These metrics are robust and reliable for evaluating generative models, offering statistical consistency under noise and perturbations. 

\subsubsection{Fidelity}
Fidelity measures how closely the generated samples resemble the real samples in the dataset. Using the TopP\&R framework, fidelity is computed based on the overlap between the estimated support of the real data and that of the generated data. This overlap is quantified through kernel density estimation (KDE) and a bootstrap-derived confidence band that ensures robustness against noise. The fidelity metric helps determine if the generated data retains key characteristics of the real data, such as constellations or signal structures, without introducing artifacts.

\subsubsection{Diversity}
Diversity evaluates whether the generative model produces outputs that span the full variability of the real data. A high diversity score indicates that the generated samples adequately represent the range of variations in the training data. Using TopP\&R, diversity is assessed by determining whether the generated samples cover the support of the real data. The robust support estimation provided by TopP\&R ensures that diversity metrics are not unduly influenced by outliers or sparsely distributed data points.

\subsubsection{Comparison with Other Metrics}
The TopP\&R framework offers several advantages over traditional metrics like Fréchet Inception Distance (FID) and Precision and Recall (P\&R). By systematically filtering out topological noise and focusing on statistically significant features, TopP\&R ensures reliable support estimation and consistent evaluation. This robustness is particularly valuable for RF signal datasets, which are prone to noise and adversarial perturbations. 

\subsection{Perceptual Analysis of Constellations}
Similarly like with VQVAE reconstructions, the perceptual method of evaluation involves plotting the I/Q constellations for each class and visually comparing them against the original constellations (Figures~\ref{fig:FakeGPT}, \ref{fig:FakeMonai}). This qualitative analysis provides an intuitive understanding of how well the reconstructions preserve class-specific signal characteristics. 
%%%%%%%%%%%%%%%%%%%%%%%%%%
\section{Results}\label{sec:res}
%%%%%%%%%%%%%%%%%%%%%%%%%%%%%
We here present performance of the proposed generative model, ReFormer, evaluated using  fidelity, diversity, and Top-F1 metrics for both the MONAI and the GPT transformer transformer. Additional quantitative metric included here is the accuracy of a pretrained classifier on the original dataset, the VQVAE reconstructions and the fakes from the both transformers. Finally, qualitative evaluation was conducted by visualizing I/Q constellations and analyzing codebook usage histograms.
%%%%%%%%%%%%%%%%%%%%%%%%%%%%%%%%%
\subsection{Quantitative Metrics}
The fidelity, diversity, and Top-F1 metrics, computed for both transformers, are summarized in Table~\ref{tab:quantitative_results}.

\begin{table}[h]
\centering
\caption{Fake signals fidelity statistics for MONAI and nano-GPT Transformers}
\label{tab:quantitative_results}
\begin{tabular}{|c|c|c|c|c|}
\hline
\textbf{Transformer} &\textbf{Parameters} & \textbf{Fidelity} & \textbf{Diversity} & \textbf{Top-F1} \\ \hline
MONAI     & 443K           & 1.0               & 0.6909             & 0.8172          \\ \hline
nano-GPT       & 36.2K           & 1.0               & 0.8455             & 0.9163          \\ \hline
\end{tabular}
\end{table}

Both transformers achieved a fidelity score of 1.0, indicating that the generated samples closely resemble the real data. However, the nano-GPT transformer outperformed the MONAI transformer in both diversity (0.8455 vs. 0.6909) and Top-F1 (0.9163 vs. 0.8172), demonstrating its ability to produce a more diverse range of samples while maintaining high accuracy.

To further validate the quality of the generated RF fakes, a pretrained classifier, initially trained on the original dataset, was tested on various datasets: the original test data, VQVAE reconstructions of the test data, and fakes generated by both the MONAI and nano-GPT transformers. The test data consisted of 500 samples per modulation class. Table~\ref{tab:classification_accuracy} summarizes these results.

\begin{table}[h]
\centering
\caption{Classification Accuracy on Different Datasets}
\label{tab:classification_accuracy}
\begin{tabular}{|l|c|}
\hline
\textbf{Dataset} & \textbf{Classification Accuracy (\%)} \\
\hline
Original Test Data & 100.00 \\
\hline
Reconstructed Test Data & 100.00 \\
\hline
GPT Generated Fake Data & 81.80 \\
\hline
MONAI Generated Fake Data & 44.07 \\
\hline
\end{tabular}
\end{table}

The classification accuracy metrics provide insight into the practical utility of the generated datasets. Both the original and reconstructed test data achieved a perfect classification accuracy of 100\%, demonstrating that the VQVAE model effectively captures and retains the essential characteristics of the RF signals, such that the classifier remains highly accurate. 

However, there is a notable difference in the performance of the fakes: the nano-GPT transformer produced fakes with a classification accuracy of 81.80\%, significantly higher than that of the MONAI transformer, which only achieved 44.07\%. This disparity underscores the nano-GPT transformer's superior capability to generate more realistic and complex RF signal fakes which maintain salient characteristics. 

The performance loss in the MONAI-generated fakes suggests potential issues in capturing the finer details of the signal characteristics, or the  overfitting to less generalizable features during the training phase (see the conclusions).

\subsection{Qualitative Evaluation}
To further assess the performance, we visualized:
\begin{itemize}
    \item \textbf{Reconstructions:} Figures~\ref{fig:FakeMonai}~and~ \ref{fig:FakeGPT} illustrate a random single-fake-based I/Q constellation for both the MONAI and GPT transformers, respectively. This includes I/Q constellations for all six modulation classes. Both generative models preserve the class-specific characteristics of the signals. The nano-GPT based fakes produced constellation diagrams that closely resemble the original, indicating effective signal learning.
    \item \textbf{Codebook Usage:} We analyzed the codebook usage across all signals for the both transformers and the VQVAE latents. The codebook usage histograms for all six classes were combined into a single figure, revealing uniform usage across all codewords, thereby validating the proper training of the VQVAE.
\end{itemize}
%%%%%%%%%%%%%%%%%%%%%%%%%%%%%%%%%%
%\subsection{Fake Samples}

\begin{figure}[h]
\centering
\includegraphics[width=0.4\textwidth,height=3.5cm]{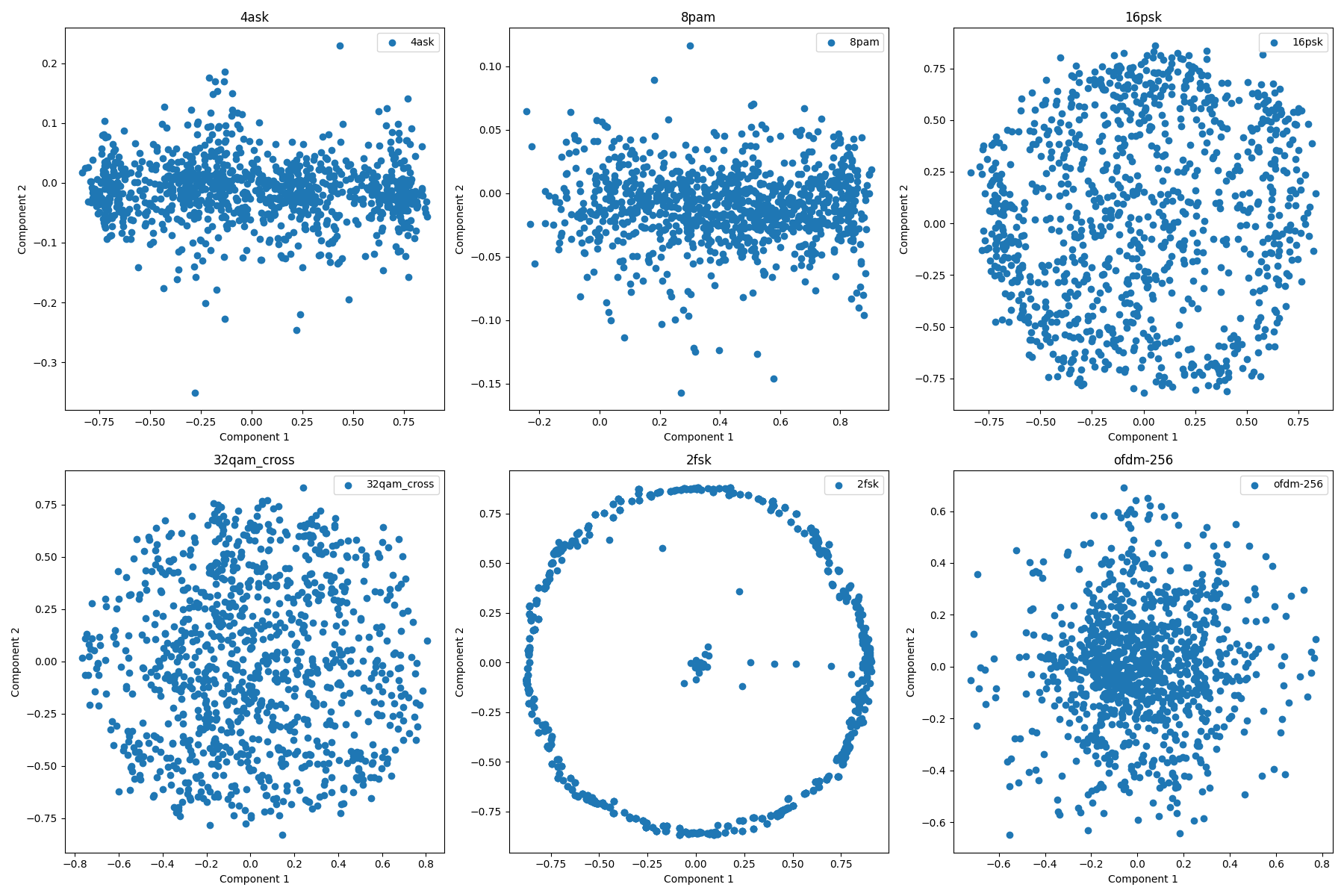} % Replace with actual file path
\caption{Reconstruction examples for GPT based transformer. The I/Q constellations for each class closely resemble the original signals.}
\label{fig:FakeGPT}
\end{figure}

\begin{figure}[h]
\centering
\includegraphics[width=0.4\textwidth,height=3.5cm]{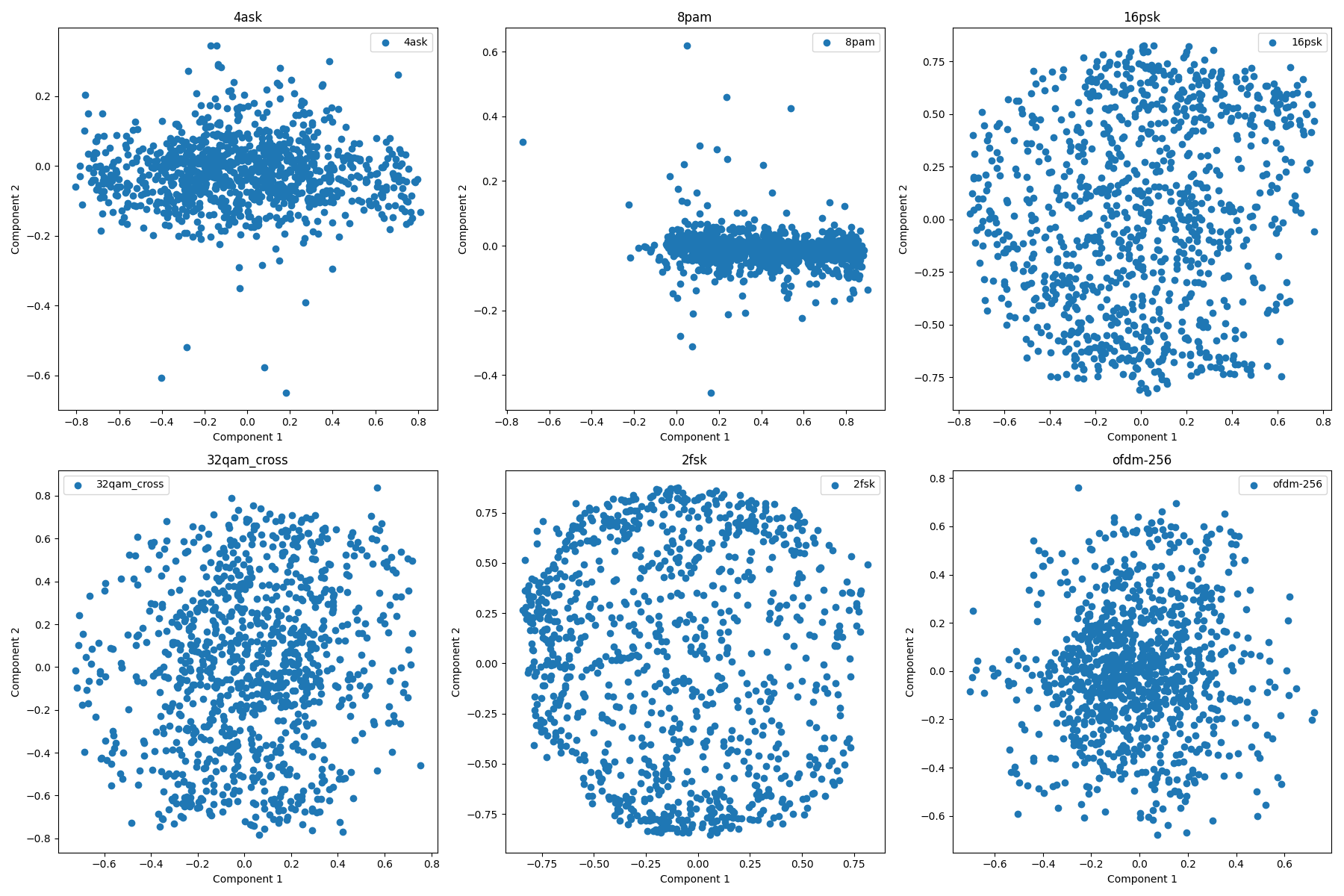} % Replace with actual file path
\caption{Reconstruction examples for MONAI transformer. GPT based Transformer creates better-quality I/Q constellations.}
\label{fig:FakeMonai}
\end{figure}
%%%%%%%%%%%%%%%%%%%%%%%%%%%
\subsection{Codebook Usage}
\begin{figure*}
\centering
\begin{subfigure}{0.2\textwidth}
    \centering
    \includegraphics[width=\textwidth]{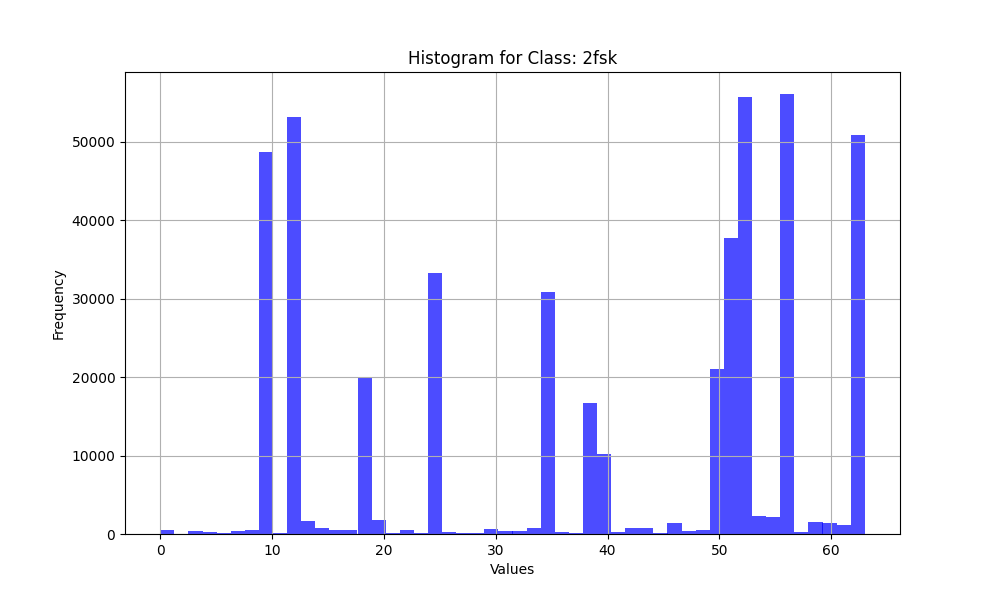}
    \caption{2FSK}
\end{subfigure}
\begin{subfigure}{0.2\textwidth}
    \centering
    \includegraphics[width=\textwidth]{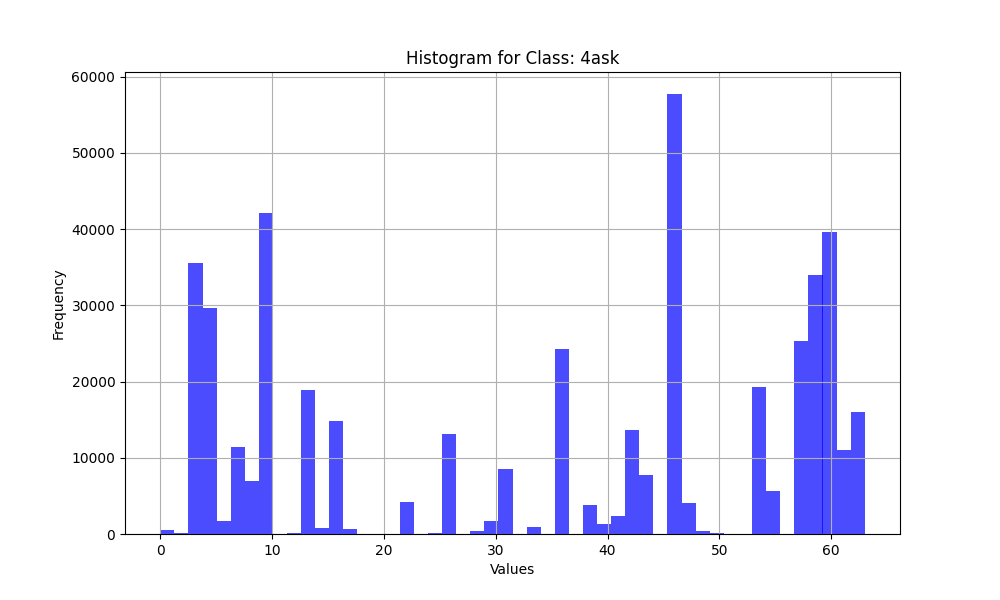}
    \caption{4ASK}
\end{subfigure}
\begin{subfigure}{0.2\textwidth}
    \centering
    \includegraphics[width=\textwidth]{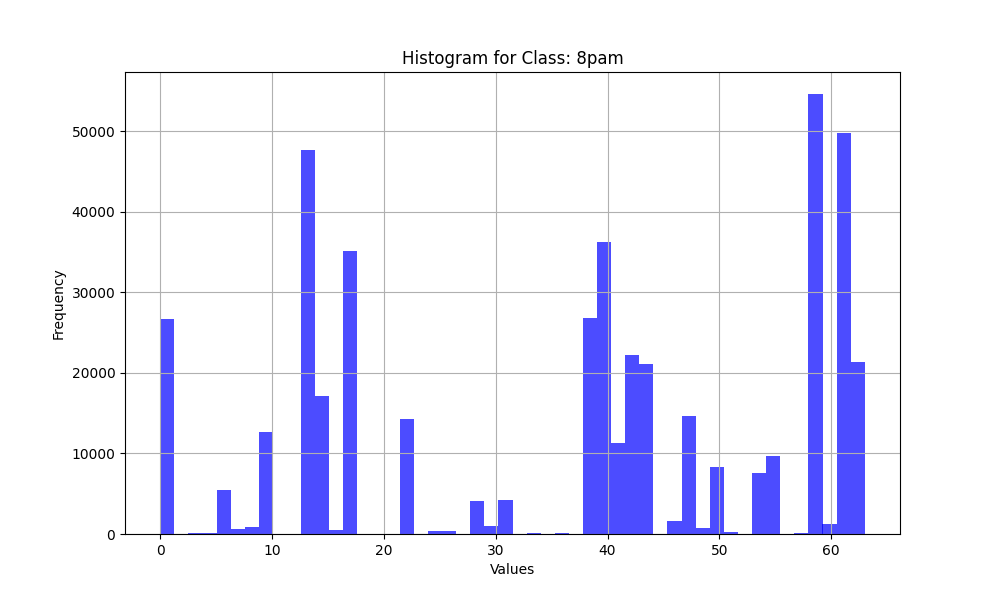}
    \caption{8PAM}
\end{subfigure}
\begin{subfigure}{0.2\textwidth}
    \centering
    \includegraphics[width=\textwidth]{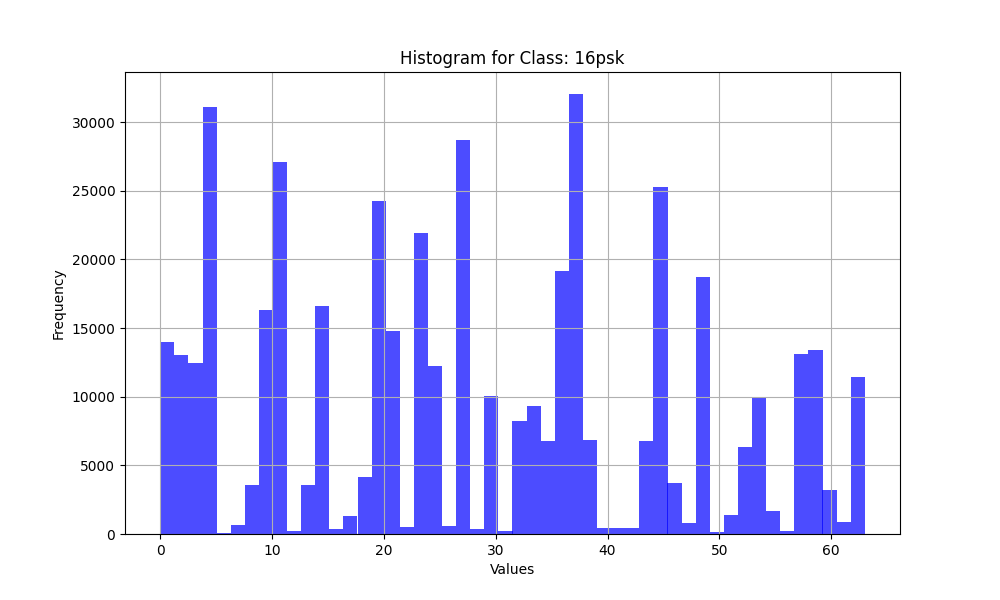}
    \caption{16PSK}
\end{subfigure}
\begin{subfigure}{0.2\textwidth}
    \centering
    \includegraphics[width=\textwidth]{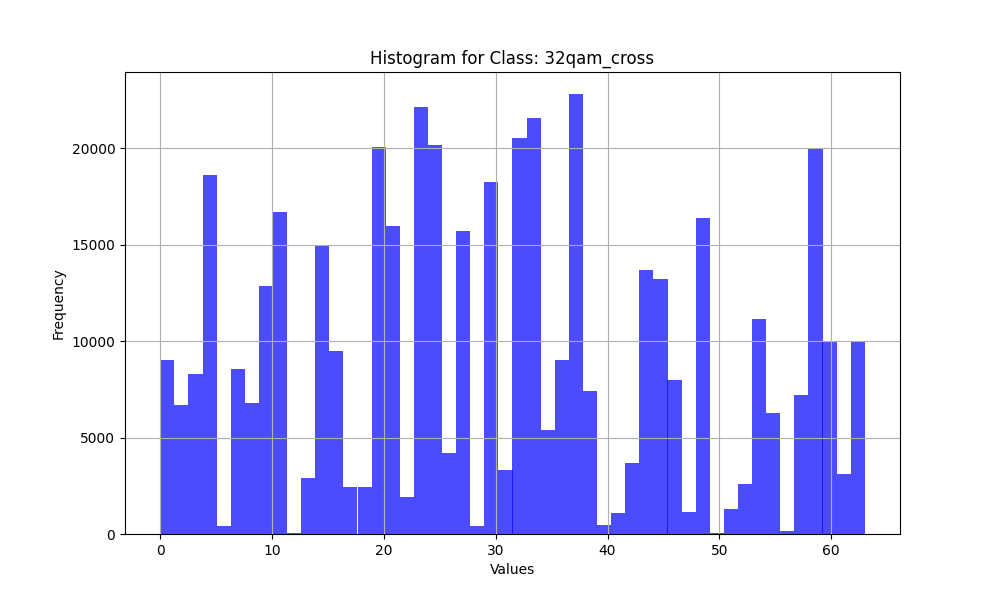}
    \caption{32QAM-Cross}
\end{subfigure}
\begin{subfigure}{0.2\textwidth}
    \centering
    \includegraphics[width=\textwidth]{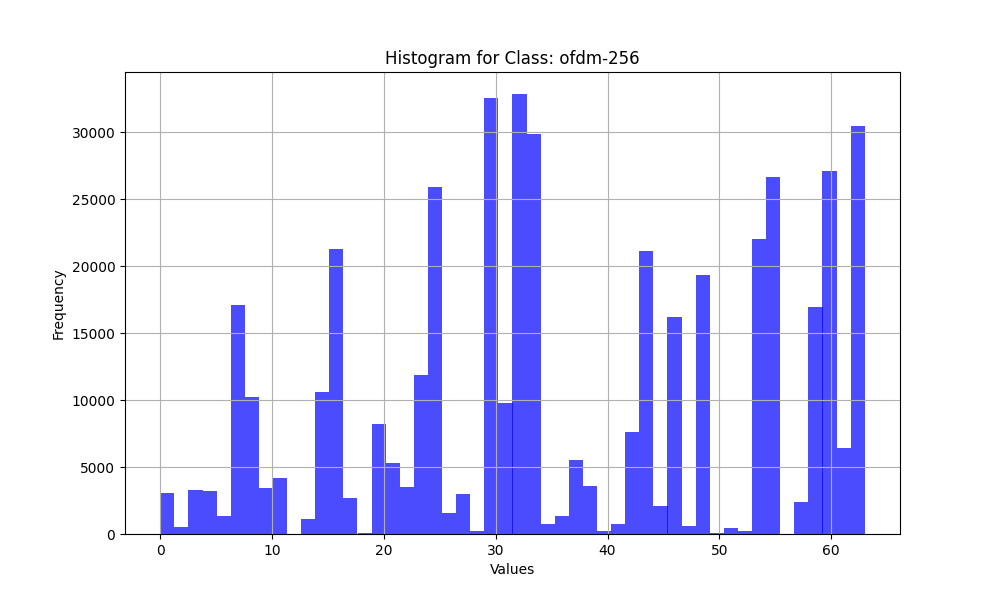}
    \caption{OFDM-256}
\end{subfigure}
\caption{Codebook usage histograms for all six modulation classes generated by the GPT transformer.}
\label{fig:codebook_usage_gpt}
\end{figure*}

Figure~\ref{fig:codebook_usage_gpt} shows the codebook usage histograms for all six modulation classes generated by the GPT transformer. Similarly, Figure \ref{fig:codebook_usage_monai} presents the codebook usage for the MONAI transformer. Each figure contains six subplots, one for each modulation class. The uniform usage of codewords across all classes indicates that the VQVAE was properly trained and did not suffer from mode collapse.

\begin{figure*}
\centering
\begin{subfigure}{0.2\textwidth}
    \centering
    \includegraphics[width=\textwidth]{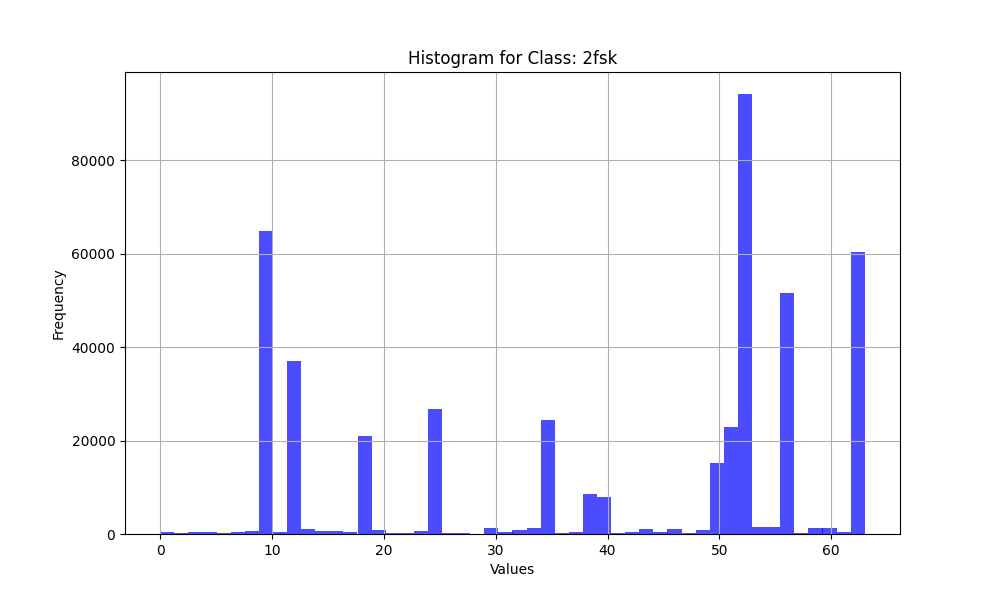}
    \caption{2FSK}
\end{subfigure}
\begin{subfigure}{0.2\textwidth}
    \centering
    \includegraphics[width=\textwidth]{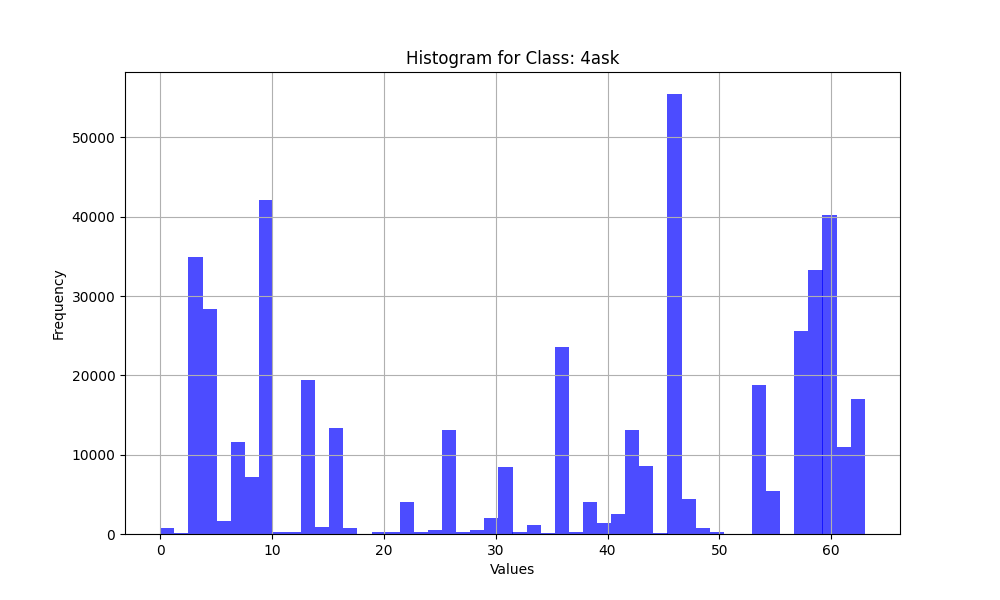}
    \caption{4ASK}
\end{subfigure}
\begin{subfigure}{0.2\textwidth}
    \centering
    \includegraphics[width=\textwidth]{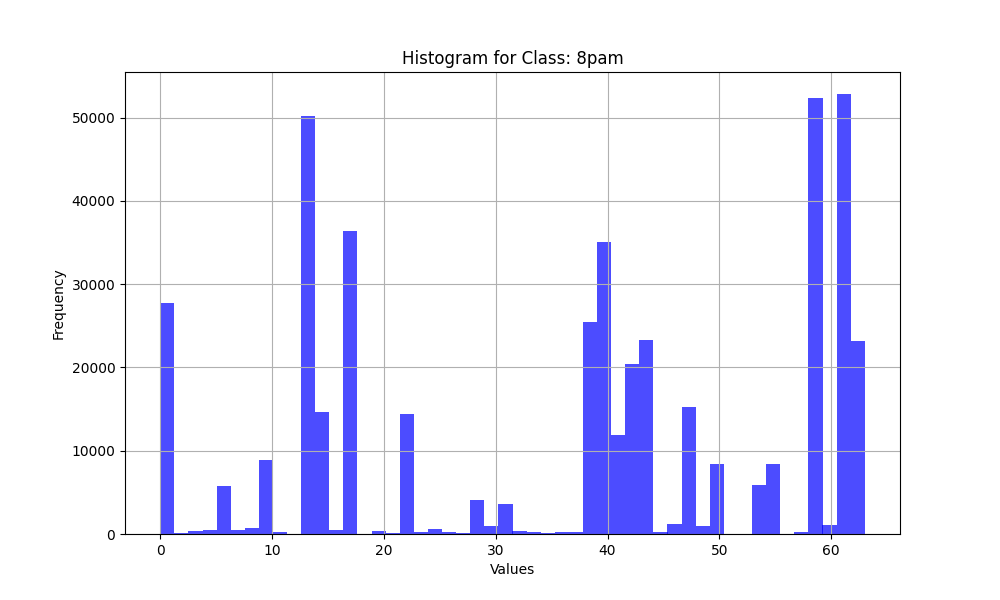}
    \caption{8PAM}
\end{subfigure}
\begin{subfigure}{0.2\textwidth}
    \centering
    \includegraphics[width=\textwidth]{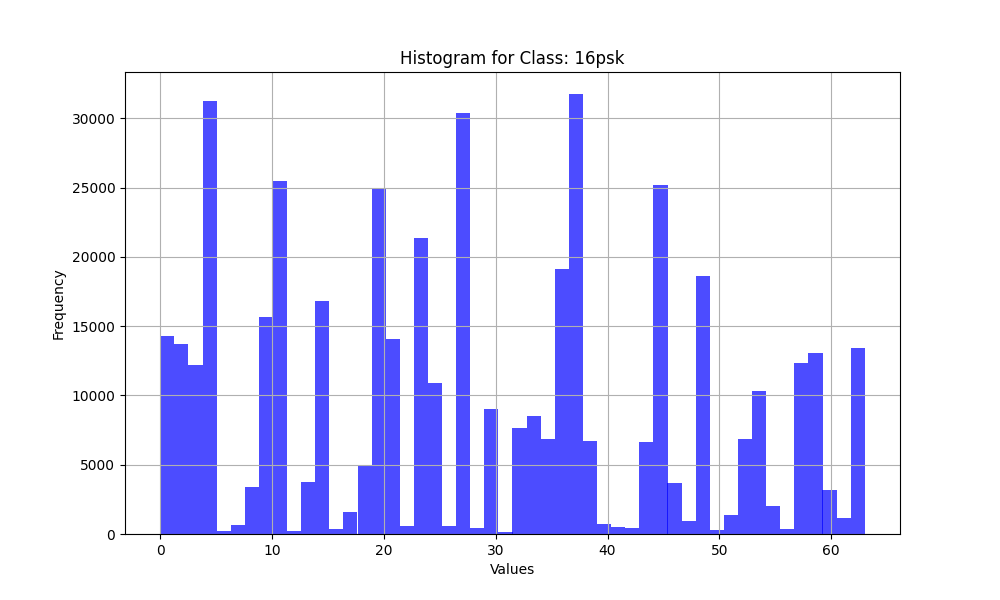}
    \caption{16PSK}
\end{subfigure}
\begin{subfigure}{0.2\textwidth}
    \centering
    \includegraphics[width=\textwidth]{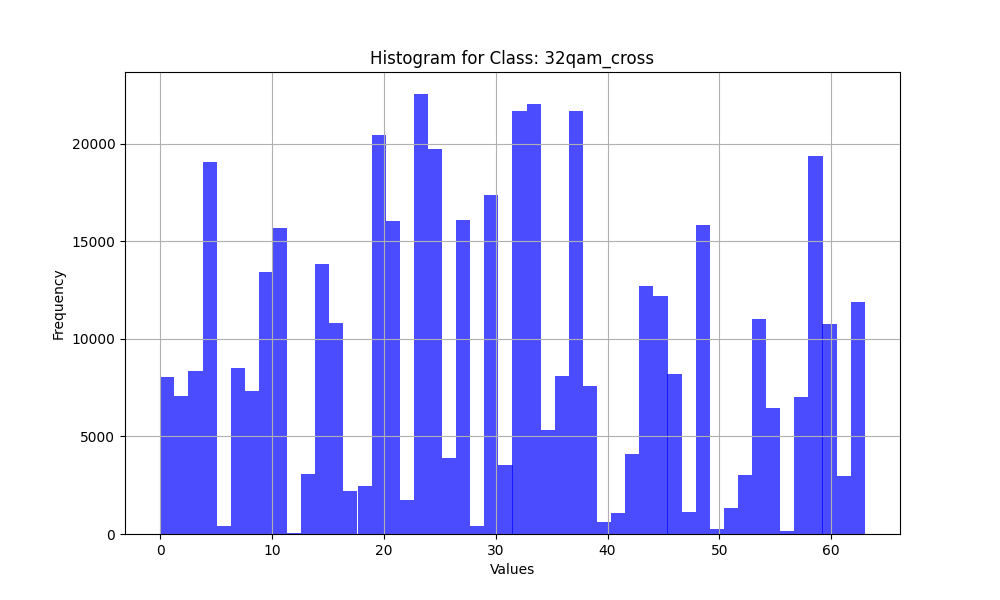}
    \caption{32QAM-Cross}
\end{subfigure}
\begin{subfigure}{0.2\textwidth}
    \centering
    \includegraphics[width=\textwidth]{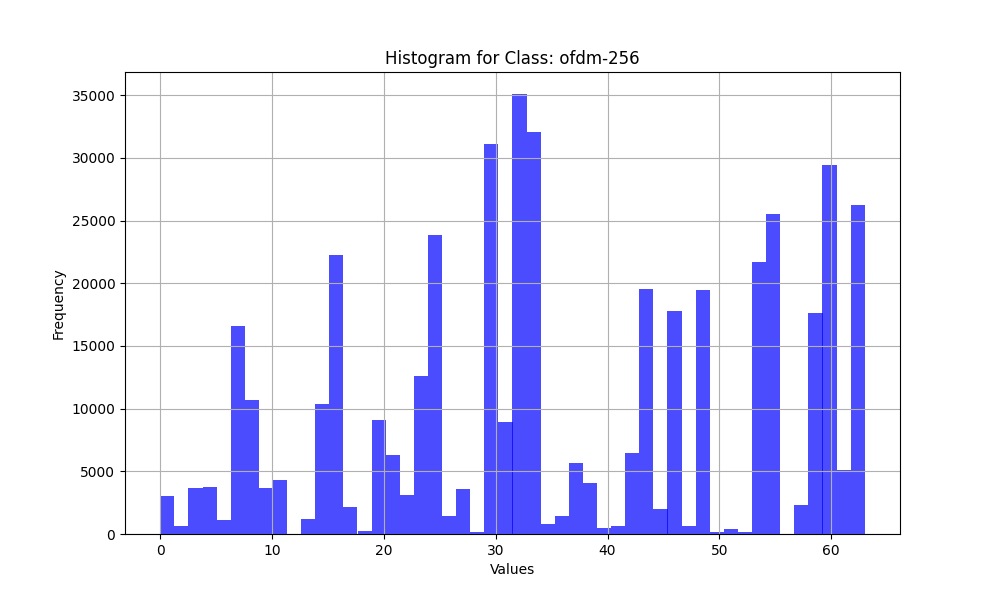}
    \caption{OFDM-256}
\end{subfigure}
\caption{Codebook usage histograms for all six modulation classes generated by the MONAI transformer.}
\label{fig:codebook_usage_monai}
\end{figure*}
%%%%%%%%%%%%%%%%%%%%%%%%%%%%%%
\subsection{Discussion}
The results indicate that both transformers are capable of generating high-fidelity RF signal fakes. While the MONAI transformer achieves satisfactory performance, the GPT transformer demonstrates superior diversity and accuracy, making it a more robust choice for generating a wide range of RF signal variations. The qualitative evaluations further support these findings, showing consistent reconstruction quality and uniform codebook usage. The fact the loss curves for the MONAI training start diverging after 100 epochs, indicate that this model is at the edge of overfitting, possibly because its number of parameters is 10 times the number of nano-GPT parameters. This may be the cause of the slightly inferior performance, both in terms of qualitative (Figures~\label{fig:FakeGPT}~and~\label{fig:FakeMonai}) and quantitative indicators (Table~\ref{tab:quantitative_results}).
%%%%%%%%%%%%%%%%%%%%%%%%%%
\section{Conclusion}
The presented results validate the effectiveness of the proposed ReFormer framework in generating RF fakes for data augmentation, with potential applications in training machine learning models for wireless communication systems. In general, the both transformer models performed surprisingly well. Even the statistics of fake tokens (codeword indices) in Figures~\ref{fig:codebook_usage_gpt},~\label{fig:codebook_usage_monai} are almost indistinguishable and similar to VQVAE statistics. This is a promising and simple approach for RF dataset augmentation.  Future work will address the optimal complexity of the  transformer model, as a function of the RF-signal dimension $p,$ the length of the discrete representation $d_s,$ and the dimension of the codebook $Q.$ Additionally, we will extend the model to include the context that can add the effects of the channel.

\vspace{-2mm}
\bibliographystyle{IEEEtran}
\bibliography{reformerbib}
\end{document}